\documentclass[lettersize,journal]{IEEEtran}

\usepackage{amsmath,amssymb,bm}
\usepackage{graphicx}
\usepackage{booktabs}
\usepackage{multirow}
\usepackage{siunitx}
\usepackage{url}
\usepackage{tabularx,array}
\usepackage{cite}
\usepackage{balance}
\usepackage{adjustbox}
\usepackage{placeins}
\usepackage{stfloats}
\usepackage{tikz}
\usetikzlibrary{arrows.meta,positioning,decorations.pathreplacing,calc}

\newenvironment{notetopractitioners}
  {\par\small\noindent\textbf{Note to Practitioners---}}
  {\par\normalsize\vspace{0.67ex}}

\title{Trajectory-Initialized Neural Double Q-Routing for Large-Scale Overhead Hoist Transport Systems}

\author{Cheng Gu, Qiusheng Zhao, Anbang Liu, Shaochong Lin, and Max Z.~J. Shen%
\thanks{This work has been submitted to the IEEE for possible publication. Copyright may be transferred without notice, after which this version may no longer be accessible.}%
\thanks{Cheng Gu, Qiusheng Zhao, Anbang Liu, and Shaochong Lin are with the Department of Data and Systems Engineering, The University of Hong Kong, Hong Kong (e-mail: chenggu03@connect.hku.hk; qszhao@hku.hk; anbang@hku.hk; shaoclin@hku.hk).}%
\thanks{Max Z.~J. Shen is with the Department of Data and Systems Engineering, The University of Hong Kong, Hong Kong, and the College of Engineering, University of California, Berkeley, CA, USA (e-mail: maxshen@hku.hk).}}

\begin{document}
\maketitle

\begin{abstract}
Large-scale industrial robot fleets share constrained physical infrastructure, making vehicle travel times dependent on safety separation, intersection access, downstream blocking, and station contention. We study this problem in overhead hoist transport (OHT) systems, a representative ceiling-mounted
material-handling system used in semiconductor fabs. Static shortest-path routing cannot account for these time-varying traffic costs, whereas tabular
Q-routing adapts online but learns each destination--node--action value independently, limiting information sharing across sparsely visited routing contexts and making startup behavior sensitive to inaccurate value estimates.
We propose Neural Double Q-routing, which replaces destination-indexed tables with a shared state--action value network. The network is warm-started through return-to-go regression on mixed simulator-generated routing trajectories and then refined online using Double-Q updates, local congestion correction, and event-stratified structured replay. Across nine matched fleet-size--arrival-rate settings with 100, 150, and 200 OHTs, the proposed framework reduces mean completion time relative to tabular Double Q-routing by $0.8\%$--$8.8\%$. It achieves the lowest mean completion time among all compared methods in the six 150- and 200-OHT settings, whereas Dijkstra remains best in the three 100-OHT settings. Completed-task counts remain within $1\%$ of tabular Double Q-routing in eight of nine settings, and 95th-percentile completion time decreases in eight settings. In two matched startup scenarios, offline initialization increases the number of completed tasks by up to $23\%$ and reduces tail completion time by up to $15\%$.
\end{abstract}

\begin{notetopractitioners}
Overhead hoist transport controllers in semiconductor factories must select routes quickly while travel conditions change because of vehicle separation, intersection access, downstream queues, blocking, and load/unload contention. Conventional shortest-path routing is straightforward to deploy but does not respond to these delays, whereas table-based adaptive routing may require
substantial online experience before its route estimates become dependable. The proposed controller changes only the next-hop decision at track split nodes; existing low-level mechanisms for vehicle separation, intersection access, and other safety functions remain in place. A compact neural value model is initialized from simulator-generated routing trajectories and subsequently updated during operation. In the evaluated simulations, its benefit depended on the operating condition: it achieved the lowest mean completion time in the 150- and 200-OHT settings, while Dijkstra remained best in the 100-OHT settings. Offline initialization also increased startup task
completions and reduced tail completion time in the two evaluated startup scenarios. Deployment requires a representative simulator, reliable local traffic measurements, and facility-specific validation of reward settings and controller interfaces. The present evidence is limited to one fab layout and one simulator; physical commissioning, initialization from operating logs, transfer to other transport systems, and robustness to layout changes remain to be evaluated.
\end{notetopractitioners}


\section{Introduction}

\begin{figure}[t]
  \centering
  \includegraphics[width=0.94\linewidth]{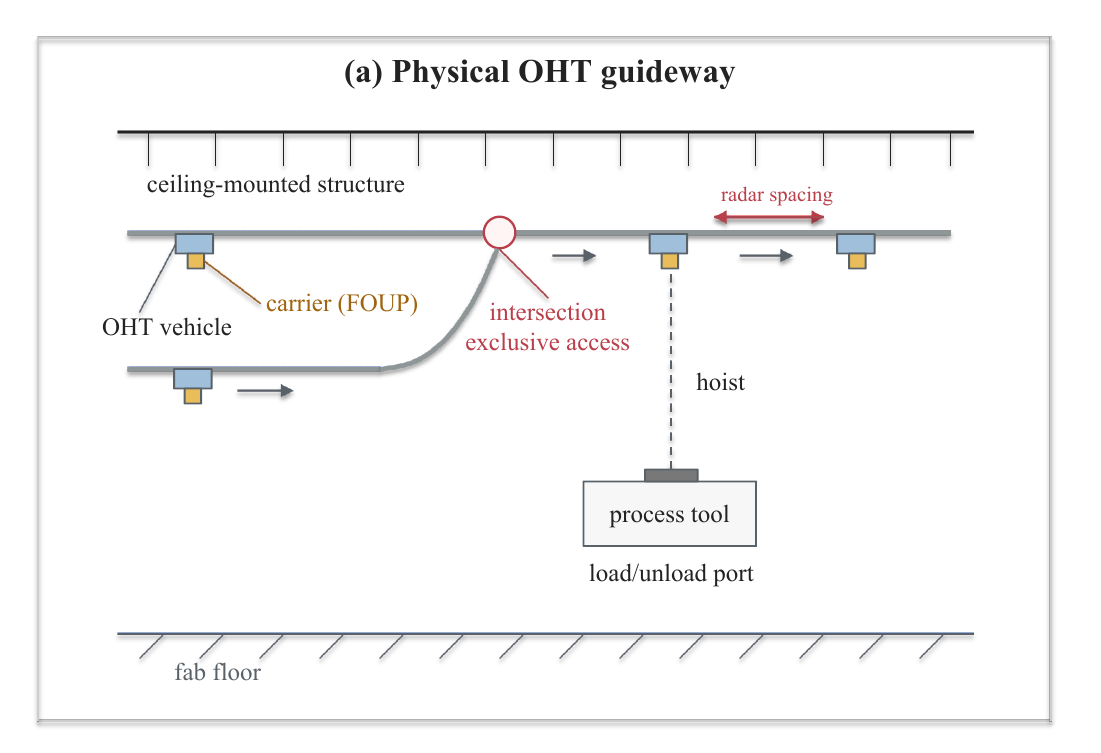}\\[-0.2ex]
  \includegraphics[width=0.92\linewidth]{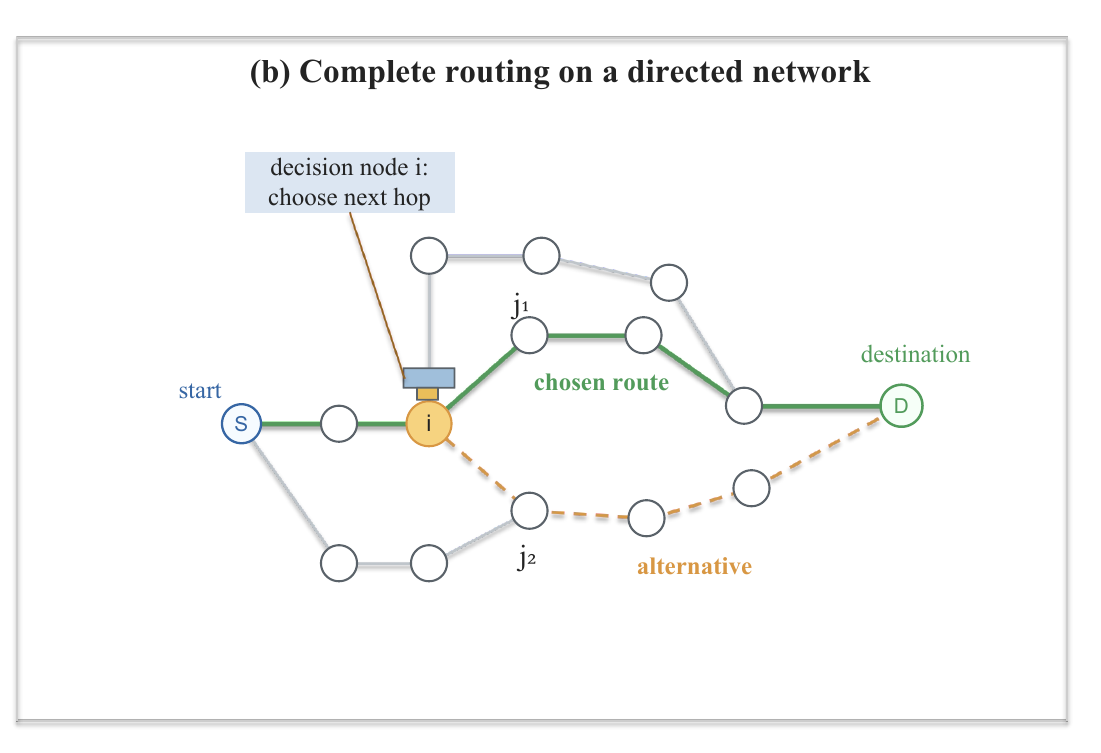}
  \caption{OHT routing context and decision process. (a) Vehicles move carriers along a ceiling-mounted, unidirectional guideway where safety spacing, shared segments, and exclusive intersections couple travel times across the fleet. (b) On the directed guideway graph, an OHT with destination $D$ makes a local next-hop decision at split node $i$, choosing between candidates such as $j_1$ and $j_2$, and repeats these decisions to form a complete route from start to destination.}
  \label{fig:oht}
\end{figure}

In a 300\,mm semiconductor fab, an automated material handling system moves front-opening unified pods among process tools, stockers, and buffers to sustain continuous production. Overhead hoist transport (OHT) vehicles execute these moves on a ceiling-mounted, unidirectional guideway (Fig.~\ref{fig:oht}). Once a transport task and its current pickup or delivery destination have been assigned, the routing controller repeatedly selects an outgoing track at each split. These choices determine when the carrier reaches its endpoint and how long the vehicle occupies shared guideway capacity. OHT routing is therefore a repeated resource-allocation decision within the fab transport schedule, while task dispatching remains separate.

The central difficulty is endogenous congestion. Safety spacing, exclusive intersection access, shared track segments, downstream admission constraints, and finite load/unload capacity couple vehicle motion~\cite{agrawal2006survey}. Consequently, the elapsed time after a route choice contains free-flow motion, waiting for separation or right-of-way, downstream blocking, queue spillback, and station contention~\cite{hwang2020q,kim2009effective}. A geometrically short route can direct a vehicle into a queue and propagate delay to following vehicles, whereas a locally longer route can avoid a contested segment. Next-hop consequences are therefore traffic-dependent and fleet-coupled. Yet each vehicle must choose from graph-feasible outgoing edges using bounded current information, while the existing low-level controller continues to enforce vehicle separation and intersection access.

A joint look-ahead controller would have to anticipate future task releases, interacting vehicle trajectories, and time-dependent resource availability across a large network. Practical OHT routing therefore commonly relies on computationally light rules, shortest paths, or dynamically adjusted edge costs~\cite{bartlett2014congestion,gupta2021improving}. Static shortest-path routing is deterministic and easy to audit, but fixed costs cannot represent current traffic. Congestion-aware routing responds to measured or predicted conditions, yet it compresses those conditions into prescribed edge penalties whose quality determines how well downstream route-choice effects are represented. We assess routing primarily by mean task completion time, while completed-task count and 95th-percentile completion time reveal unfinished work or delay shifted into the tail.

Q-routing addresses the downstream-effect problem by learning the expected remaining travel time after a vehicle at node $i$ selects neighbor $j$ toward destination $d$~\cite{boyan1993packet,hwang2020q}. Its destination-conditioned table supports fast local decisions and adapts from realized transitions. However, every destination--node--neighbor tuple $(d,i,j)$ has an independent value, so one update cannot directly inform another tuple with similar route progress and local traffic. On the evaluated guideway, $608$ destinations and $4257$ directed edges require approximately $2.59$ million tabular Q entries, and tabular Double Q-routing doubles that storage. Frequently visited tuples may stabilize while rarely visited tuples remain data-poor. A shared candidate-edge scorer instead updates one function used across nodes and destinations, although parameter sharing alone does not establish value accuracy. We retain Q-routing's local feasible-action interface but learn discounted return under a composite routing reward rather than remaining travel time alone.

\begin{figure}[h]
  \centering
  \includegraphics[width=0.95\columnwidth]{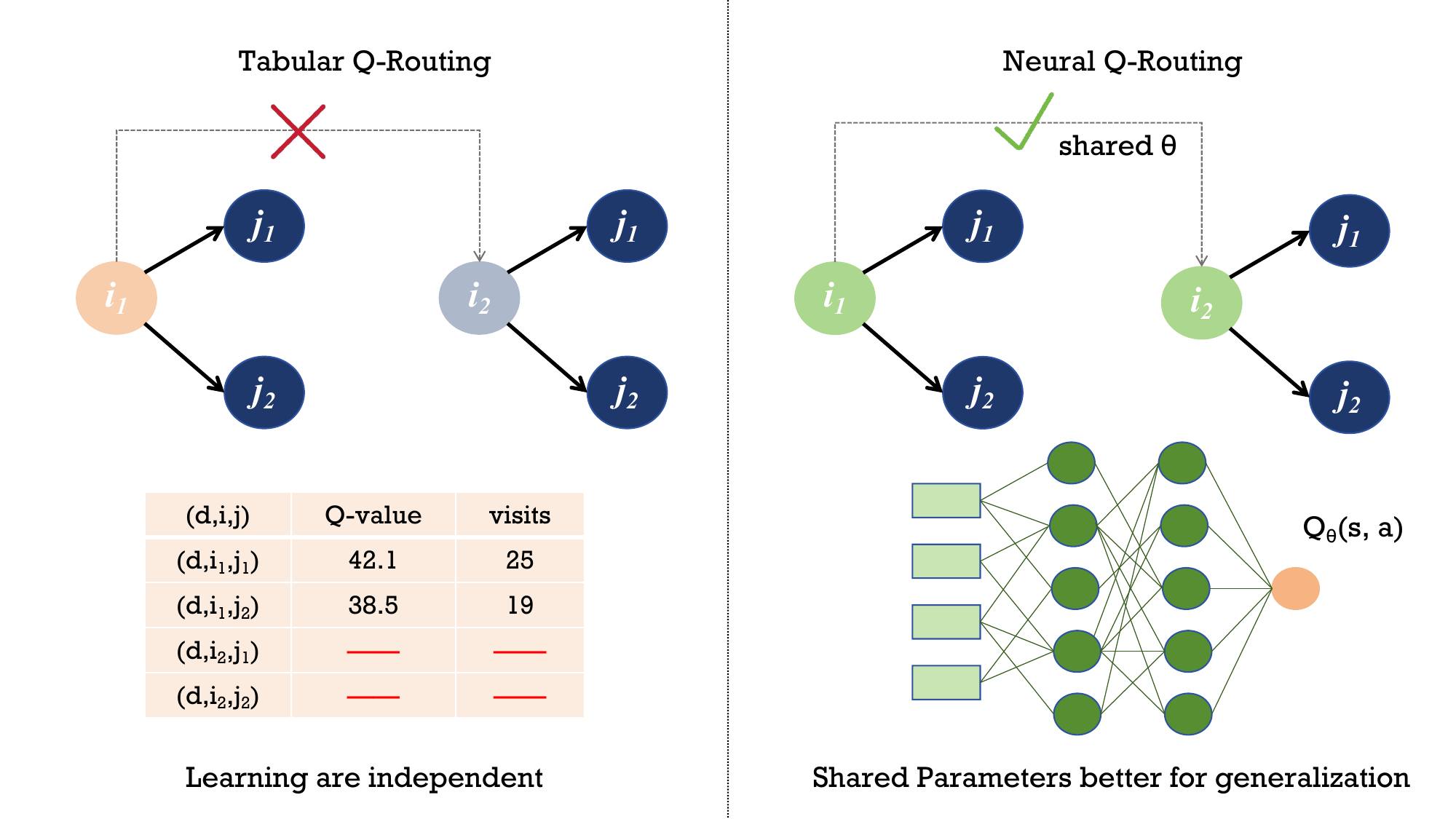}
  \caption{Tabular versus neural Q-routing. Tabular Q-routing stores an independent value for each $(d,i,j)$ tuple, whereas a neural value function $Q_\theta(s,a)$ scores candidate actions from a common state--action feature vector $\phi(s,a)$. Shared parameters let observations from frequently visited contexts inform the fitted scoring rule used at less frequently visited contexts.}
  \label{fig:tabular_vs_neural}
\end{figure}

Replacing the table with a neural function addresses only the representation problem. A continuing online router must also be useful before accumulating extensive new experience and must adapt without becoming unstable or unresponsive. Random initialization provides no informative startup ranking, and using one noisy estimator for action selection and evaluation can introduce positive bootstrap bias~\cite{hasselt2010double}. A fleet-wide value model can also lag behind short-lived local queues, while uniform replay may underrepresent infrequent blocking, near-deadlock, and load/unload bottlenecks. The design problem is therefore to combine a compact shared representation with informative initialization and stable yet responsive online refinement.

We propose Neural Double Q-routing, an offline-warm-started, online-adaptive next-hop router for fab-scale OHT systems. A shared neural network scores each controller-admitted edge from task, topology, route-progress, and bounded local-traffic features, while transitions from all vehicles update the same value function. Before operation, supervised return-to-go regression on simulator-generated Dijkstra, Q-routing, and Double Q-routing trajectories supplies an initial estimate. Online, Double-Q targets separate action selection from evaluation, recent edge-level residuals correct local rankings without entering the recursive bootstrap, and event-stratified structured replay retains transitions from predefined congestion and service-event classes. The model changes only split-node route selection; low-level safety controls remain outside the value function.

We evaluate the framework in one discrete-event simulator on a fab guideway with $3684$ nodes, $4257$ directed edges, and $608$ destinations. Nine matched settings combine $100$, $150$, and $200$ OHTs with arrival rates of $1.0$, $1.5$, and $2.0$ task\,s$^{-1}$. Relative to tabular Double Q-routing, Neural Double Q-routing lowers mean completion time in all nine settings by $0.8\%$--$8.8\%$; completed-task counts remain within $1\%$ in eight, and 95th-percentile completion time decreases in eight. It obtains the lowest mean among all methods in the six 150- and 200-OHT settings, whereas Dijkstra remains best in the three 100-OHT settings. The evidence therefore indicates a setting-dependent benefit rather than universal superiority. In matched startup scenes, offline initialization changes early mean completion time modestly but increases completed tasks by $22.2\%$--$23.1\%$ and reduces tail completion time by $7.2\%$--$15.0\%$.

The contributions are as follows:
\begin{enumerate}
\item \textbf{A compact shared value representation for repeated OHT routing.} A candidate-edge function replaces independent destination--node--neighbor entries while retaining the local graph-feasible action interface. Its parameter count is independent of $|D|$, $|V|$, and $|E|$, reducing $2.59$ million tabular Q entries to $5825$ neural parameters in the evaluated system.
\item \textbf{An offline-to-online procedure for startup, stable refinement, and local responsiveness.} Mixed trajectories initialize the shared value function; Double-Q targets refine global estimates; local residuals adjust recent edge rankings; and event-stratified replay retains infrequent congestion evidence.
\item \textbf{A matched-scene evaluation of gains and setting-dependent behavior.} A fleet-size and arrival-rate sweep compares static and tabular alternatives, while warm-start and component studies separate startup effects from the online-refinement mechanisms.
\end{enumerate}

The remainder of this paper is organized as follows. Section~\ref{sec:related} reviews OHT control layers, congestion-aware and learning-based routing, deep traffic representations, and offline-to-online value learning. Section~\ref{sec:problem} formulates the event-driven OHT routing problem, including the guideway model, state--action features, decision-interval dynamics, reward, and evaluation criteria. Section~\ref{sec:method} presents the shared neural value function, offline warm start, online Double-Q refinement, local congestion correction, and structured replay. Section~\ref{sec:experiments} describes the matched-scene protocol and reports the operating-regime sweep and ablation studies. Section~\ref{sec:discussion} interprets the findings and states their limitations, and Section~\ref{sec:conclusion} concludes the paper. Appendix~\ref{sec:appendix} provides the complexity analysis, full feature definitions, value-estimate stability analysis, and implementation details.

\section{Related Work}
\label{sec:related}

This section positions Neural Double Q-routing along five related lines of research. We first distinguish route guidance from the neighboring OHT decisions of system design, dispatching, and rebalancing. We then review congestion-aware path construction and learning-based traffic prediction. Next, we trace the development from tabular Q-routing to neural, destination-conditioned value approximation. We subsequently discuss trajectory-based initialization, offline-to-online learning, and nonuniform experience replay. Finally, we locate the proposed method with respect to these lines in terms of its decision variable, representation, data regime, and online adaptation mechanism.

\subsection{Routing Within OHT Control}

Manufacturing transport control spans decisions made at different temporal and spatial scales. At the production-system level, integrated scheduling methods may jointly select operations, machines, and transport resources. For
example, Zhang et al.~\cite{zhang2023dynamic} use heterogeneous graph representations and deep reinforcement learning for flexible-job-shop
scheduling with insufficient vehicles. Such formulations coordinate production and transport decisions rather than isolating vehicle routing.

OHT control in a semiconductor fab can itself be separated into system design, vehicle dispatching, idle-capacity rebalancing, and route guidance.
Layout and simulation studies evaluate guideway capacity, traffic mechanisms, and system-level performance under alternative designs \cite{agrawal2006survey, wang2004performance}. Dispatching assigns an
outstanding transport request to a vehicle. Hungarian assignment, prediction-based dispatching, and vehicle look-ahead formulations use current or anticipated requests to reduce pickup and delivery delay \cite{kim2009effective, wan2022predictive, benzoni2023vehicle}. Rebalancing
instead relocates idle vehicles toward areas with anticipated demand. Graph-based multi-agent policies have been developed for zone rebalancing \cite{ahn2021cooperative} and coordinated bay-level vehicle allocation \cite{chou2025multiagent}.

These decisions interact because dispatching and rebalancing alter the traffic encountered by the routing controller. Nevertheless, their immediate control
variables differ. Dispatching determines which vehicle serves a request, rebalancing determines where idle capacity waits, and route guidance
determines how a vehicle with an assigned task proceeds through the guideway. The present work fixes the task-dispatching rule and studies the repeated selection of a controller-admitted outgoing edge after a vehicle has a current task and destination.

\subsection{Congestion-Aware Path Construction and Traffic Prediction}

A major family of dynamic OHT routing methods retains a conventional shortest-path procedure while replacing static edge lengths with traffic-dependent costs. Bartlett et al.~\cite{bartlett2014congestion} update a computationally light congestion measure and reroute vehicles as measured
conditions change. Gupta et al.~\cite{gupta2021improving} estimate
congestion-dependent edge-traversal costs for Dijkstra routing and study routing costs together with pickup rules. Machine-learning traffic control has likewise been used to predict congestion at critical bottlenecks and
redirect vehicles through modified routing costs \cite{lee2020intelligent}. In these approaches, measured or predicted traffic is translated into edge weights before a path-construction algorithm is applied.

More recent methods learn richer travel-time or traffic representations. Ahn et al.~\cite{ahn2022congestion} represent successive OHT traffic states as
graphs and use a graph-convolutional gated recurrent model to predict future edge-travel times for congestion-aware rerouting. Choi et
al.~\cite{choi2024dynamic} combine a local heuristic approximation for the near portion of a candidate route with a deep neural approximation of its more distant travel time. Separable contextual graph neural networks identify
spatiotemporal tailgating-oriented congestion patterns but do not themselves
produce routing actions~\cite{lee2024separable}. HarmonyRouting instead uses a graph neural network to estimate route-level traffic impact and incorporates
that prediction into system-level route selection
\cite{kang2026harmonyrouting}. These methods exploit graph-wide or route-wide information to predict quantities subsequently used by a routing procedure.

Reinforcement learning has also been used to influence routes through intermediate traffic-control variables. Kang et al.~\cite{kang2019dynamic}
identify heavily used and frequently congested regions and use deep reinforcement learning to adjust the routing cost imposed on those regions.
Koo et al.~\cite{koo2026alleviation} partition a fab-scale guideway into areas and learn dynamic link-weight adjustments that influence downstream
path selection. At a broader system level, Lee et
al.~\cite{lee2025autonomous} combine active Q-routing, discrete-event simulation, and real-time digital-twin information in an OHT orchestration framework.

The common feature of these approaches is that learning supplies a traffic prediction, path estimate, or network-level cost intervention. The routing
algorithm then constructs or modifies a path using that intermediate quantity. Neural Double Q-routing exposes a different control interface: at
each split, it directly evaluates the controller-admitted outgoing edges and selects one next hop from the current task, route-progress, topology, and bounded local-traffic context.

\subsection{From Tabular Q-Routing to Neural Route Guidance}

Q-routing was introduced for adaptive packet forwarding in dynamically changing communication networks. It stores a destination-conditioned estimate
$Q(d,i,j)$ of the remaining delivery time when an item at node $i$ first
moves to neighbor $j$ on its way to destination $d$
\cite{boyan1993packet}. The value is updated from realized local delay and a downstream estimate, yielding an inexpensive online next-hop lookup. The
original study also considered neural approximation of the routing value but reported inconclusive results, so replacing the table with a function
approximator was already recognized as a possible extension. Predictive
Q-routing subsequently retained favorable past estimates and modeled traffic
recovery to improve adaptation when network load changes
\cite{choi1996predictive}.

Q-routing has been adapted to semiconductor OHT systems because its local decision epoch and low online computational cost fit guideway control. Hwang and Jang~\cite{hwang2020q} develop a $Q(\lambda)$-based dynamic route-guidance method with eligibility traces and
exploration mechanisms. Hong et al.~\cite{hong2022practical} further study Q-learning-based route guidance together with vehicle assignment in
practical-scale OHT settings. These methods learn from observed OHT traffic while retaining a distinct value for each destination--node--neighbor tuple.
Consequently, experience obtained at one tuple does not directly update another tuple with similar route progress or congestion conditions.

Neural value approximation provides a way to share statistical structure
across routing contexts. In communication networks, You et
al.~\cite{you2022packet} embed deep value networks in multi-agent Q-routing, allowing individual routers to estimate next-hop values from destination and
richer traffic context. More generally, DQN represents action values with a deep neural network and stabilizes their learning through replay and target networks~\cite{mnih2015human}. Double Q-learning separates action selection from action evaluation to reduce maximization bias \cite{hasselt2010double}, and Double DQN extends this separation to neural
function approximation~\cite{van2016deep}. Within OHT routing, Ao et al.~\cite{ao2024dynamic} encode the track map, vehicle state, and task information as a raster and train a convolutional map-information Double-DQN planner.

Conditioning one value function on multiple destinations is also related to
universal value function approximation. Universal value function approximators condition a shared value model on both the current state and a goal, enabling parameters to be shared across goal-dependent control problems
\cite{schaul2015universal}. This provides the general conceptual basis for a
destination-conditioned shared value function. It does not, by itself, determine the routing decision interface, the available traffic information, or the manner in which the function should be initialized and updated in an OHT system.

Accordingly, the present work does not treat neural approximation, destination conditioning, or the Double-DQN target as individually new.
Relative to per-node deep Q-routing and map-wide convolutional planning, its
representation uses one fleet-wide state--action function to score each candidate outgoing edge from a fixed-dimensional feature vector. The same parameters are used across split nodes, destinations, vehicles, and task
phases, while the existing controller continues to determine which outgoing edges are feasible. This parameter sharing is evaluated on the fixed fab
guideway and should not be interpreted as zero-shot generalization to unseen layouts or destinations.

\subsection{Trajectory-Based Initialization and Online Adaptation}

Previously collected control trajectories can be used to improve the initial behavior of an online value learner. Deep Q-learning from demonstrations
pretrains a Q-network from demonstration data and continues learning online using temporal-difference, supervised action-classification, and prioritized replay objectives~\cite{hester2018dqfd}. Its objective encourages the learned policy to reproduce demonstrated actions while retaining the ability to improve beyond the demonstrator. The present offline stage has a different role: it regresses return-to-go targets from trajectories generated by multiple routing policies and does not impose an imitation loss favoring the actions of a single demonstrator.

Offline reinforcement learning addresses the stronger problem of optimizing a policy or value function from a fixed dataset while controlling errors on poorly represented actions. Conservative value estimation and behavior-regularized policy improvement are representative approaches
\cite{kumar2020conservative,fujimoto2021minimalist}. Moving such a policy online introduces an additional distribution transition as newly collected experience begins to replace or supplement the offline data. Balanced replay with pessimistic estimates, policy expansion, calibrated value initialization, and periodic policy revitalization have been proposed to manage this transition in general control domains
\cite{lee2022balanced,zhang2023policy,nakamoto2023calql,kong2024efficient}.

Neural Double Q-routing does not optimize a separate conservative offline
policy. Its simulator-generated trajectories provide supervised return-to-go targets for initializing the same compact value function that is subsequently refined by online Double-DQN updates. The trajectories are
collected from Dijkstra, Q-routing, and Double Q-routing so that the prior is not tied to a single behavior policy. Thus, the offline component is better viewed as trajectory-based value initialization than as a complete offline-RL stage.

Experience replay determines which online observations continue to influence the value model. Prioritized experience replay ranks individual transitions using temporal-difference error~\cite{schaul2015prioritized}, whereas selective replay has considered criteria such as surprise, reward,
distribution matching, and state-space coverage
\cite{isele2018selective}. Event-stratified structured replay uses a domain-specific criterion instead: transitions are assigned to mutually exclusive OHT event classes, and update batches use predefined class quotas and loss weights. It therefore emphasizes observed congestion and service conditions rather than ranking samples by their current TD error, and it is not presented as an importance-sampling correction.

Online adaptation also requires responding to traffic changes that may be visible initially at only a small number of edges. Earlier predictive Q-routing modifies route estimates through stored favorable experience and
predicted traffic recovery~\cite{choi1996predictive}. In the proposed controller, the shared neural value remains the recursively learned base estimate, while a recent edge- and task-phase-specific TD residual adjusts the
current action ranking outside the bootstrap target. A count-dependent term is used only during the online warmup period. This separation allows recent local observations to affect immediate decisions without recursively propagating the local correction through the shared value function.

\subsection{Positioning of Neural Double Q-Routing}

The closest methods can be distinguished along four dimensions. First, dispatching and rebalancing choose vehicles or idle-capacity locations, traffic predictors and link-weight controllers modify inputs to path construction, and route-guidance methods choose a next hop for an already assigned vehicle. Second, tabular Q-routing stores separate
destination--node--neighbor values, whereas map- and graph-based methods use global spatial representations; the proposed model instead scores one candidate edge from bounded state--action features using parameters shared throughout the fixed guideway. Third, offline RL learns a policy from a fixed dataset, whereas the proposed offline data are used only to initialize the value function that continues to learn online. Fourth, generic prioritized replay ranks transitions by learning signals, whereas the proposed replay mechanism retains predefined OHT congestion and service-event classes.

Neural Double Q-routing therefore combines established reinforcement-learning components within an OHT-specific deployment architecture: a local controller-compatible next-hop decision, a shared candidate-edge value representation, mixed-trajectory initialization, Double-DQN refinement, and online mechanisms for recent and infrequent congestion evidence. This combination---rather than neural approximation, Double Q-learning, or offline pretraining in isolation---defines the methodological position of the work.

\section{OHT Routing Problem Formulation}
\label{sec:problem}

This section formalizes the OHT routing problem in three parts. First, we describe the directed guideway and define the split-based routing process.
Second, we explain how learning is shared across the fleet while routing decisions are executed vehicle-wise, and specify the state and candidate-action representations used at each decision epoch. 
Finally, we characterize the variable-duration decision intervals, define the routing reward, and present the criteria used to evaluate controller performance.

\subsection{Guideway Model and Split-Based Routing Process}

\begin{figure}[!t]
\centering
\includegraphics[width=\columnwidth]{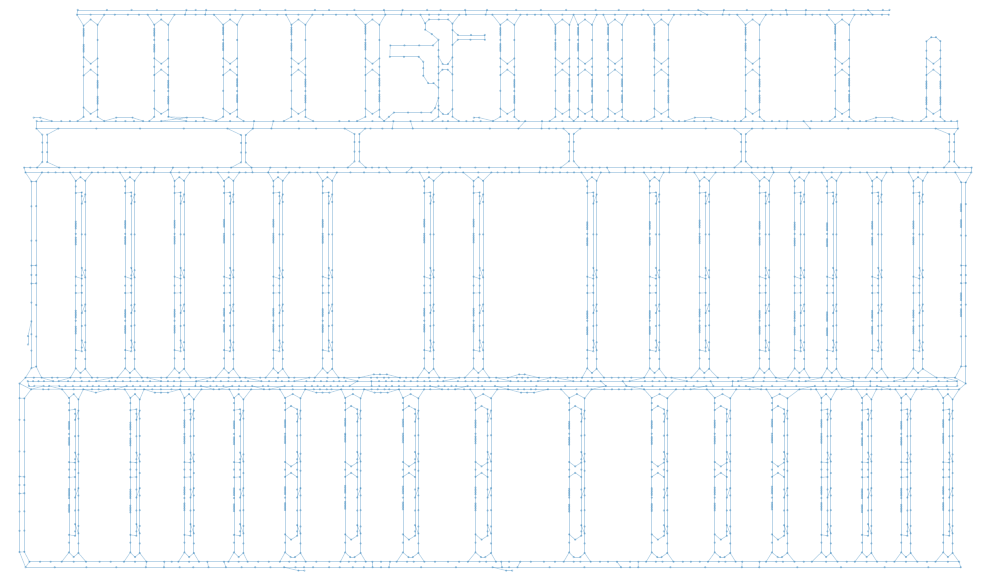}
\begingroup
\makeatletter
\long\def\@makecaption#1#2{%
  \@IEEEfigurecaptionsepspace
  \parbox[t]{\hsize}{\normalfont\footnotesize\centering
    #1.\nobreakspace\nobreakspace #2\par}}
\makeatother
\caption{Guideway graph used in the simulation. Small markers indicate control nodes, including the degree-one nodes that preserve track geometry.}
\label{fig:fab_map}
\endgroup
\end{figure}

We model the OHT guideway as a directed graph $G=(V,E)$, where nodes represent load/unload access points, intersections, and geometric control points, and directed
edges represent unidirectional track segments. The full map used in our experiments (Fig.~\ref{fig:fab_map}) contains $|V|=3684$ nodes, $|E|=4257$ directed edges, and $|D|=608$ distinct destinations. Its maximum out-degree is
$\deg^+_{\max}=2$, meaning that every split is binary. A destination is a
task-relevant endpoint node in $V$ at which a vehicle either picks up or drops
off a carrier. In practice, these destinations correspond to load/unload ports at process tools, stockers, and buffers. The destination set $D\subset V$ is therefore determined by the fab layout rather than by the task stream.

Routing decisions are made at vehicle-specific split events. A decision occurs when a vehicle reaches a split node $i$ satisfying $\deg^+(i)\!\ge\!2$, and the
corresponding action selects one outgoing edge. After the action is selected, the vehicle follows any intervening degree-one segments until it reaches the next split or until the current task phase terminates at its pickup or delivery endpoint. Consequently, different routing decisions may span different numbers of physical edges and different amounts of elapsed time. We therefore model the
routing process as semi-Markov~\cite{bradtke1994reinforcement}.

Learning operates on the embedded sequence of routing decision epochs. Each routing choice contributes one discount step, whereas the realized holding time is incorporated into the interval reward defined in Eq.~\eqref{eq:reward}. The resulting value function ranks candidate actions under this decision-epoch discounted routing criterion.

\subsection{Fleet-Wide Learning and Decision Representation}

Learning is shared across the fleet. Transitions collected from all vehicles enter common replay storage and are used to update a single value function $Q_\theta$. During execution, each vehicle applies this shared function to its controller-admitted outgoing edges using its own task state and bounded local traffic summaries. The resulting architecture combines centralized data aggregation and parameter learning with vehicle-wise local next-hop selection.

\textbf{State and candidate-action features.}
At a split node $i$ with target $d$, the state vector $s\in\mathbb{R}^{d_s}$, where $d_s=10$, contains the normalized indices of $i$ and $d$, the normalized shortest-path distance $h(i,d)$, the in- and out-degrees of $i$, a load/unload-node indicator, a three-component task-phase code representing pickup, delivery, or other, and the vehicle's recent congestion-waiting time.

The available actions are the graph out-neighbors of $i$. For each candidate
next hop $j$, the action vector
$\phi_a(s,j)\in\mathbb{R}^{d_a}$, where $d_a=14$, contains the normalized
candidate index, edge length, remaining distance $h(j,d)$, signed progress
$h(i,d)-h(j,d)$, candidate in- and out-degrees, edge occupancy,
candidate-node queue length, and six local pressure summaries. These six
summaries comprise bottleneck pressure, spillback risk, maximum and mean
one-hop downstream pressure, maximum two-hop pressure, and the fraction of
vehicles on the candidate edge whose waiting flag is active. All continuous
components are scaled and clipped to $[0,1]$. Appendix~\ref{app:features}
specifies each feature dimension and its normalization.

\textbf{Local pressure summaries.}
More precisely, let $c_{ij}$ denote the normalized occupancy of candidate edge
$(i,j)$, $q_j$ the normalized queue at node $j$,
$p_{\max}^{(1)}$ and $p_{\mathrm{mean}}^{(1)}$ the maximum and mean
occupancies of the edges leaving $j$, and $p_{\max}^{(2)}$ the maximum
occupancy one additional hop downstream. The six local pressure summaries are
\begin{equation}
\begin{split}
&\max\{c_{ij},p_{\max}^{(1)},p_{\max}^{(2)}\},\quad
\min\{1,(q_j+p_{\max}^{(1)})/2\},\\
&p_{\max}^{(1)},\quad p_{\mathrm{mean}}^{(1)},\quad
p_{\max}^{(2)},\quad
n_{ij}^{\mathrm{wait}}/\max\{1,n_{ij}\},
\end{split}
\label{eq:pressure_features}
\end{equation}
where $n_{ij}$ counts the vehicles on edge $(i,j)$ and
$n_{ij}^{\mathrm{wait}}$ counts those with an active waiting flag. Empty
downstream sets contribute zero.

The features are assembled from the vehicle's task state, the static map, and
the simulator's current edge-occupancy and node-queue records. Computing the
one- and two-hop summaries requires information only from this bounded local
neighborhood. Parameter updates, in contrast, aggregate transitions collected
from the entire fleet.

\subsection{Decision-Interval Dynamics and Routing Criterion}

\textbf{Decision-interval dynamics.}
Let $\mathcal{I}_t$ denote the ordered sequence of physical edges traversed
during decision interval $t$. The elapsed time of the interval is
\begin{equation}
    \tau_t = \sum_{e\in\mathcal{I}_t}
    \left(m_{t,e}+w_{t,e}+b_{t,e}\right),
    \label{eq:tau}
\end{equation}
where $m_{t,e}$ is the time spent moving on edge $e$, $w_{t,e}$ is the time
spent waiting for separation or right-of-way control, and $b_{t,e}$ is the
time blocked by downstream conditions or resource-admission constraints. For
an individual vehicle, $w_{t,e}$ and $b_{t,e}$ depend on the motion and
service states of the rest of the fleet. The transition to $s_{t+1}$ is
therefore stochastic even when the map and selected candidate edge are fixed.

\textbf{Routing reward.}
The reward accumulates the realized time and deadlock warnings over the
physical edges traversed during the interval. Risk and route progress are
evaluated once for the action selected at the initial split:
\begin{equation}
\begin{split}
r_t ={}&-\!\sum_{e\in\mathcal{I}_t}\!\left(
w_m m_{t,e}+w_w w_{t,e}+w_b b_{t,e}+w_d\omega_{t,e}\right)\\
&-w_r\rho(s_t,a_t)+w_p\delta_h(s_t,a_t),
\end{split}
\label{eq:reward}
\end{equation}
where $\rho(s_t,a_t)$ is the maximum of the selected-edge occupancy,
candidate-node queue, bottleneck pressure, spillback risk, and maximum one-hop
downstream pressure. The term $\omega_{t,e}$ is the deadlock-warning level
recorded on physical edge $e$. Route progress is defined as
\[
\delta_h(s_t,a_t)
=
\operatorname{clip}
\left(
\frac{h(i,d)-h(j,d)}{D_s},-1,1
\right),
\]
where $D_s$ is the distance scale. The corresponding action feature is mapped affinely to $[0,1]$, whereas the reward retains the signed progress value.

\textbf{Routing criterion and parameters.}
The reward defines a domain-weighted routing criterion. Its time-based terms retain physical units while assigning different costs to movement, waiting,
and blocking. The risk, warning, and progress terms provide anticipatory signals for congestion and route advancement. These shaping terms are not
constructed as potential differences~\cite{ng1999policy}; instead, they are components of the learned routing criterion rather than a policy-invariant transformation of elapsed time.

We fix
\[
(w_m,w_w,w_b,w_r,w_d,w_p)
=
(1.0,0.8,1.2,0.1,0.15,0.3)
\]
and use warning levels $\omega\in\{0,1,3\}$ in every cell.
Appendix~\ref{app:impl_reward} provides the remaining threshold definitions.

\textbf{Evaluation criteria.}
Controller performance over a wall-clock horizon of $T=1000$\,s is evaluated using mean completion time (CT Mean). We additionally report the number of completed tasks (Completed) and the 95th-percentile completion time (CT P95) to identify policies that either leave tasks unfinished or shift delays into the tail of the completion-time distribution. Startup analyses use an additional early-horizon window.

The resulting formulation specifies the decision process, input representation, transition dynamics, and routing criterion used by the offline-to-online Neural Double Q-routing method developed in Section~\ref{sec:method}.


\section{Offline-to-Online Neural Double Q-Routing}
\label{sec:method}
\begin{figure*}[!t]
\centering
\includegraphics[width=0.60\textwidth]{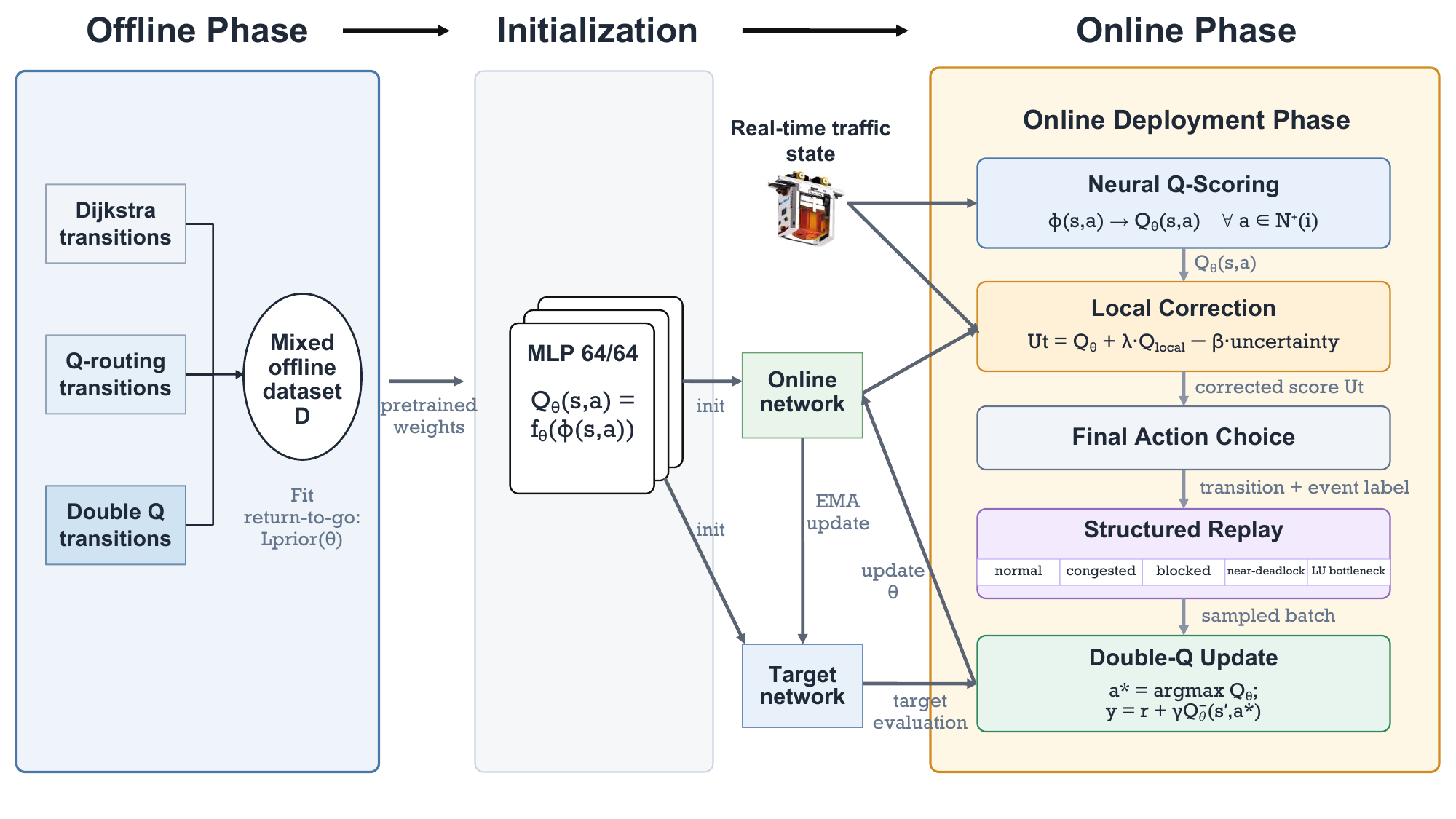}
\caption{Method overview. Logged trajectories initialize the online and target value networks. During operation, Double-Q updates refine the shared value function, while a local congestion correction adjusts action rankings from recent edge-level observations and event-stratified structured replay retains predefined congestion and service-event classes.}
\label{fig:method_overview}
\end{figure*}

This section develops Neural Double Q-routing in four parts. We first define the shared state--action neural value function and then describe its offline warm start from mixed routing trajectories. Next, we present the online Double-Q refinement rule. Finally, we detail the local congestion correction and event-stratified structured replay used during online adaptation. Fig.~\ref{fig:method_overview} summarizes the complete flow from offline initialization to online action selection and learning. We refer to the complete method as \emph{Neural Double Q-routing}; tables and figures use the compact implementation label \emph{QNeuralDouble}.

\subsection{Shared State--Action Neural Value Function}

Tabular Q-routing stores an independent value $Q(d,i,j)$ for every destination-node-action tuple. When the guideway has thousands of nodes and hundreds of destinations, most tuples are visited infrequently, and information learned at one tuple does not transfer to structurally similar situations elsewhere on the map.

We replace this table with a neural value function that scores each candidate next hop through a shared parameterization. For a vehicle in state $s$ considering action $a$, we construct a joint feature vector $\phi(s,a)$ and predict:
\begin{equation}
Q_{\theta}(s,a) = f_{\theta}\bigl(\phi(s,a)\bigr),
\label{eq:nn_q}
\end{equation}

where $f_{\theta}$ is a two-hidden-layer MLP with $64$ units per layer and ReLU activations. The state ($d_s=10$) and action ($d_a=14$) features are exactly those defined in Section~\ref{sec:problem}, giving joint input dimension $d_\phi=24$; all inputs are pre-normalized to comparable scales. The scalar output approximates the discounted return-to-go under the reward in Eq.~\eqref{eq:reward}; actions with larger values are therefore preferred. At each junction the router first evaluates $Q_{\theta}(s,a)$ for every feasible next hop. Section~\ref{sec:online_adaptation} defines the local correction applied to these base values before the final action is selected. Including biases, the MLP has 5,825 trainable parameters versus $\sim$2.59M entries in tabular Q on our guideway (Appendix~\ref{app:complexity}).

Because $\theta$ is shared across nodes and destinations, every update changes
one scoring function used throughout the map rather than one isolated table
entry. Appendix~\ref{app:param_sharing} examines the resulting estimate
stability; it does not treat lower variance as evidence of value accuracy.

\subsection{Offline Return-to-Go Warm Start}

Random initialization leaves the value model uninformative at the start of
operation. We instead pretrain $f_\theta$ on logged decision-interval
trajectories generated by Dijkstra, Q-routing, and Double Q-routing.

Each record contains $(s_t,a_t,r_t,s_{t+1},z_t)$, where $z_t$ marks a terminal
task-phase boundary. Within a trajectory segment ending at index $T_\ell$, the
regression target is
\begin{equation}
\hat G_t=\sum_{k=t}^{T_\ell}\gamma^{k-t}r_k.
\label{eq:rtg}
\end{equation}
The accumulation is reset at changes in episode, vehicle, task, target, task
phase, or terminal segment, so rewards from distinct routing episodes are not
combined. We fit the prior by minimizing
\begin{equation}
\mathcal{L}_{\mathrm{prior}}(\theta) = \mathbb{E}_{(s,a)\sim\mathcal{D}}\!\left[\bigl(Q_\theta(s,a) - \hat{G}(s,a)\bigr)^2\right].
\label{eq:offline_loss}
\end{equation}
The learned weights initialize both online $Q_\theta$ and target
$Q_{\bar\theta}$. Training batches balance the logged sources by operating
cell, behavior policy, and collection seed so that a large stratum does not
dominate the regression. Each of the three behavior policies is collected with
three seeds. The prior is trained for 40 epochs with Adam at learning rate
$10^{-3}$ and batch size 64; Appendix~\ref{app:impl_offline} gives the dataset
construction details.

\subsection{Online Double-Q Refinement}
\label{sec:online_doubleq}

Standard Q-learning's $\max_{a'} Q_{\theta}(s',a')$ bootstrap systematically overestimates noisy values---amplified in OHT routing where candidate hops often have similar true values and sparse visit counts. We use $\gamma=0.99$ on the embedded decision chain described in Section~\ref{sec:problem}; interval duration is already represented in $r$ through Eq.~\eqref{eq:reward}. The Double-DQN form of the Double-Q idea~\cite{hasselt2010double,van2016deep} separates action selection from value evaluation:
\begin{equation}
a^{\star} = \arg\max_{a'} Q_{\theta}(s',a'), \qquad
y = r + \gamma(1-z) Q_{\bar\theta}(s', a^{\star}),
\label{eq:doubleq}
\end{equation}
where $z=1$ for a terminal transition; in that case $y=r$ and no next action
is evaluated.
The online network is updated by minimizing the Huber loss:
\begin{equation}
\mathcal{L}_{\mathrm{td}}(\theta) = \mathbb{E}\!\left[\ell_{\mathrm{Huber}}\!\bigl(Q_{\theta}(s,a) - y\bigr)\right].
\label{eq:td_loss}
\end{equation}
Target networks stabilize DQN-style learning~\cite{mnih2015human,van2016deep}.
Here $Q_{\bar\theta}$ is updated by Polyak averaging
$\bar\theta\leftarrow(1-\tau)\bar\theta+\tau\theta$ with $\tau=0.005$.
Online updates use Adam at learning rate $10^{-3}$, batch size 64, and
gradient-norm clipping at 5. Updates begin after 64 replay records and are
performed every four observed transitions.

\subsection{Congestion-Aware Online Adaptation}
\label{sec:online_adaptation}

\textbf{Local congestion correction.} The shared value model can lag behind a
short-lived local bottleneck. We therefore combine its raw return estimate with
a local TD-residual estimate and a count-based uncertainty penalty:
\begin{align}
U_t(s,a) ={}& Q_{\theta}(s,a)
 + \lambda_t\,\kappa(s,a)\,R_{\mathrm{local}}(s,a) \notag\\
&{}- \beta_t\,(n(s,a)+1)^{-1/2},
\label{eq:local_correction}
\end{align}
where $R_{\mathrm{local}}$ is an exponentially averaged, clipped TD residual in
the same reward units as $Q_\theta$, and $\kappa(s,a)$ downweights an edge-level
residual carried across task phases. Here $n(s,a)$ is the observation count for
the corresponding directed edge and task phase; an edge-level fallback uses
the count pooled across phases. The schedules $\lambda_t$ and $\beta_t$
increase the role of observed local residuals and remove the initial
uncertainty penalty over $N_{\mathrm{warmup}}$ transitions. For candidate edge
and phase pair $(e,p)$, the residual update is
\begin{equation}
R_{e,p}\leftarrow(1-\alpha)R_{e,p}+
\alpha\,\operatorname{clip}\!\left(y_t-Q_\theta(s,a),-c,c\right),
\label{eq:local_residual_update}
\end{equation}
with $\alpha=0.05$ and $c=5$. We set $\kappa=1$ for an exact edge--phase
record, $0.5$ for an edge-level record pooled across phases, and $0$ when no
record exists. The schedules are
\begin{equation}
\lambda_t=0.5\min(t/5000,1),\qquad
\beta_t=0.3\max(1-t/5000,0).
\label{eq:local_schedules}
\end{equation}
The executed action is
$a_t=\arg\max_{a\in\mathcal{A}(s_t)}U_t(s_t,a)$. The two quantities have
separate roles: $Q_\theta$ is the recursively trained base return estimator,
whereas $U_t$ is the action score used by the behavior policy at execution.
Transitions generated by that policy enter replay, and $Q_\theta$ is updated
off-policy using the base Double-Q target in Eq.~\eqref{eq:doubleq}. Keeping the
edge-level residual outside the bootstrap lets recent observations alter the
current action ranking without recursively propagating those residuals through
future targets.

\textbf{Event-stratified structured replay (SR).} Uniform sampling can make
transitions involving blocking or prolonged waiting uncommon in an update
batch. We assign each transition exactly one of five labels---normal,
congested, blocked, near-deadlock, or load/unload (LU) bottleneck---using an
explicit severity order. Non-normal transitions are also retained in a recent
event pool. Each update draws a configured share from this pool, allocates that
share across event labels, and applies fixed class-dependent loss weights. In
the reported configuration, event records target $40\%$ of a batch, giving
nominal whole-batch shares of $16/12/8/4\%$ for congested, blocked,
near-deadlock, and LU-bottleneck records; normal records fill the remaining
$60\%$. This is domain-informed oversampling with a cost-sensitive loss, not
an importance sampling correction. Unlike PER~\cite{schaul2015prioritized}, which ranks
individual samples by TD error, the event labels are determined from observed
OHT traffic and service conditions. Overlapping conditions are resolved by
\begin{equation}
\begin{split}
\text{near-deadlock}&\succ\text{LU-bottleneck}\succ\text{blocked}\\
&\succ\text{congested}\succ\text{normal}.
\end{split}
\label{eq:event_priority}
\end{equation}
Table~\ref{tab:replay_labels} gives the triggers and loss weights; missing
event classes are filled first from other event records and then from general
replay.

\begin{table}[t]
\centering
\caption{Structured-replay labels and nominal whole-batch shares when all classes are available. Weight multiplies TD loss.}
\label{tab:replay_labels}
\scriptsize
\setlength{\tabcolsep}{2pt}
\begin{tabularx}{\columnwidth}{@{}lXcc@{}}
\toprule
Label & Trigger & Share & Weight \\
\midrule
normal & otherwise & $60\%$ & $1.0$ \\
congested & $w_t \ge 1.5\,$s & $16\%$ & $1.5$ \\
blocked & $b_t > 0$ & $12\%$ & $2.0$ \\
near-deadlock & $\omega_t = 3$ & $8\%$ & $3.0$ \\
LU-bottleneck & at load/unload node and $w_t > 15\,$s & $4\%$ & $2.0$ \\
\bottomrule
\end{tabularx}
\end{table}


\section{Experimental Evaluation}
\label{sec:experiments}

This section first specifies the simulation setup, baselines, and matched-scene protocol. It then reports the long-horizon results across the fleet-size and arrival-rate sweep, and finally presents the warm-start and online-component ablations. Within each operating cell, every method receives the same task sequence and initial conditions.

\begin{table*}[t]
\centering
\caption{Matched-scene \SI{1000}{s} long-horizon results across the fleet $\times$ arrival-rate sweep. Values are mean $\pm$ std over 5 seeds.}
\label{tab:main_sweep}
\scriptsize
\setlength{\tabcolsep}{3.5pt}
\begin{adjustbox}{max width=\textwidth}
\begin{tabular}{rr | rrrr | rr | rrr}
\toprule
 & & \multicolumn{4}{c|}{CT Mean (s)} & \multicolumn{2}{c|}{$\Delta$ CT Mean of Ours} & \multicolumn{3}{c}{Other metrics of Ours} \\
OHT & AR & D & Q & QDouble & \textbf{Ours} & vs QDouble (s) & $\Delta$ \% & Compl. & CT P95 (s) & $\Delta$ P95 \\
\midrule
100 & 1.0 & \textbf{319.84 $\pm$ 3.46} & 326.41 $\pm$ 4.31 & 324.18 $\pm$ 3.21 & 321.06 $\pm$ 2.86 & $-$3.12 & $-$1.0\% & 363.0 $\pm$ 3.2 & 594.21 $\pm$ 9.50 & $-$0.3\% \\
100 & 1.5 & \textbf{374.62 $\pm$ 4.17} & 379.95 $\pm$ 4.94 & 382.47 $\pm$ 3.83 & 377.86 $\pm$ 3.32 & $-$4.61 & $-$1.2\% & 385.0 $\pm$ 2.6 & 712.88 $\pm$ 11.36 & $+$0.2\% \\
100 & 2.0 & \textbf{413.05 $\pm$ 4.55} & 416.78 $\pm$ 5.47 & 418.04 $\pm$ 4.23 & 414.27 $\pm$ 3.67 & $-$3.77 & $-$0.9\% & 374.0 $\pm$ 4.2 & 765.40 $\pm$ 12.30 & $-$0.5\% \\
\midrule
150 & 1.0 & 205.91 $\pm$ 4.88 & 197.18 $\pm$ 4.37 & 195.06 $\pm$ 3.54 & \textbf{177.84 $\pm$ 2.07} & $-$17.22 & $-$8.8\% & 588.0 $\pm$ 8.1 & 362.77 $\pm$ 8.30 & $-$6.2\% \\
150 & 1.5 & 187.30 $\pm$ 4.50 & 176.85 $\pm$ 3.89 & 174.42 $\pm$ 3.22 & \textbf{168.50 $\pm$ 2.12} & $-$5.92 & $-$3.4\% & 916.2 $\pm$ 12.7 & 337.04 $\pm$ 7.75 & $-$3.9\% \\
150 & 2.0 & 200.04 $\pm$ 4.75 & 188.27 $\pm$ 4.12 & 185.69 $\pm$ 3.40 & \textbf{174.86 $\pm$ 2.23} & $-$10.83 & $-$5.8\% & 1182.2 $\pm$ 16.5 & 364.18 $\pm$ 8.44 & $-$3.4\% \\
\midrule
200 & 1.0 & 121.27 $\pm$ 1.97 & 120.43 $\pm$ 2.02 & 118.74 $\pm$ 1.66 & \textbf{117.20 $\pm$ 1.54} & $-$1.54 & $-$1.3\% & 544.0 $\pm$ 5.4 & 263.18 $\pm$ 4.99 & $-$0.6\% \\
200 & 1.5 & 141.18 $\pm$ 2.54 & 138.42 $\pm$ 2.27 & 136.85 $\pm$ 1.88 & \textbf{135.71 $\pm$ 1.50} & $-$1.14 & $-$0.8\% & 847.8 $\pm$ 8.4 & 281.93 $\pm$ 5.41 & $-$0.9\% \\
200 & 2.0 & 235.86 $\pm$ 4.27 & 223.40 $\pm$ 3.74 & 213.79 $\pm$ 3.04 & \textbf{203.62 $\pm$ 2.17} & $-$10.17 & $-$4.8\% & 1045.0 $\pm$ 10.3 & 376.61 $\pm$ 7.19 & $-$1.3\% \\
\bottomrule
\end{tabular}
\end{adjustbox}
\end{table*}

\subsection{Setup}
\label{sec:setup}

All methods are evaluated on the fab guideway of Section~\ref{sec:problem} ($|V|\!=\!3684$, $|E|\!=\!4257$, $|D|\!=\!608$) under a common discrete-event simulator. We sweep fleet size $\in\{100,150,200\}$ and task arrival rate (AR) $\in\{1.0,1.5,2.0\}$ task\,s$^{-1}$, giving $3{\times}3$ cells. Within each cell every method shares the same task stream, initial vehicle placement, and random seed. The simulation horizon is $T=1000$\,s, and startup analyses use the first $200$\,s. We report mean completion time (CT Mean), completed tasks, and 95th-percentile completion time (CT P95). In the warm-start study, Early CT is the mean CT of tasks completed within the first $200$\,s.

\textbf{Baselines.} \textbf{Dijkstra} ignores dynamics; tabular \textbf{Q-routing}~\cite{boyan1993packet,hwang2020q} adapts to congestion without bias correction; tabular \textbf{Double Q-routing}~\cite{hasselt2010double} adds overestimation control. \textbf{QNeuralDouble} (ours) uses the full framework: offline prior, neural value function, online Double-Q updates, local correction, and structured replay.

\subsection{Long-horizon results across the sweep}
\label{sec:longhorizon}

Table~\ref{tab:main_sweep} reports matched-scene results over the $1000$\,s horizon for all nine cells. Bold marks the best method per cell; $\Delta$ is Ours vs.\ QDouble (negative is improvement).

\begin{figure*}[t]
\centering
\includegraphics[width=0.81\textwidth]{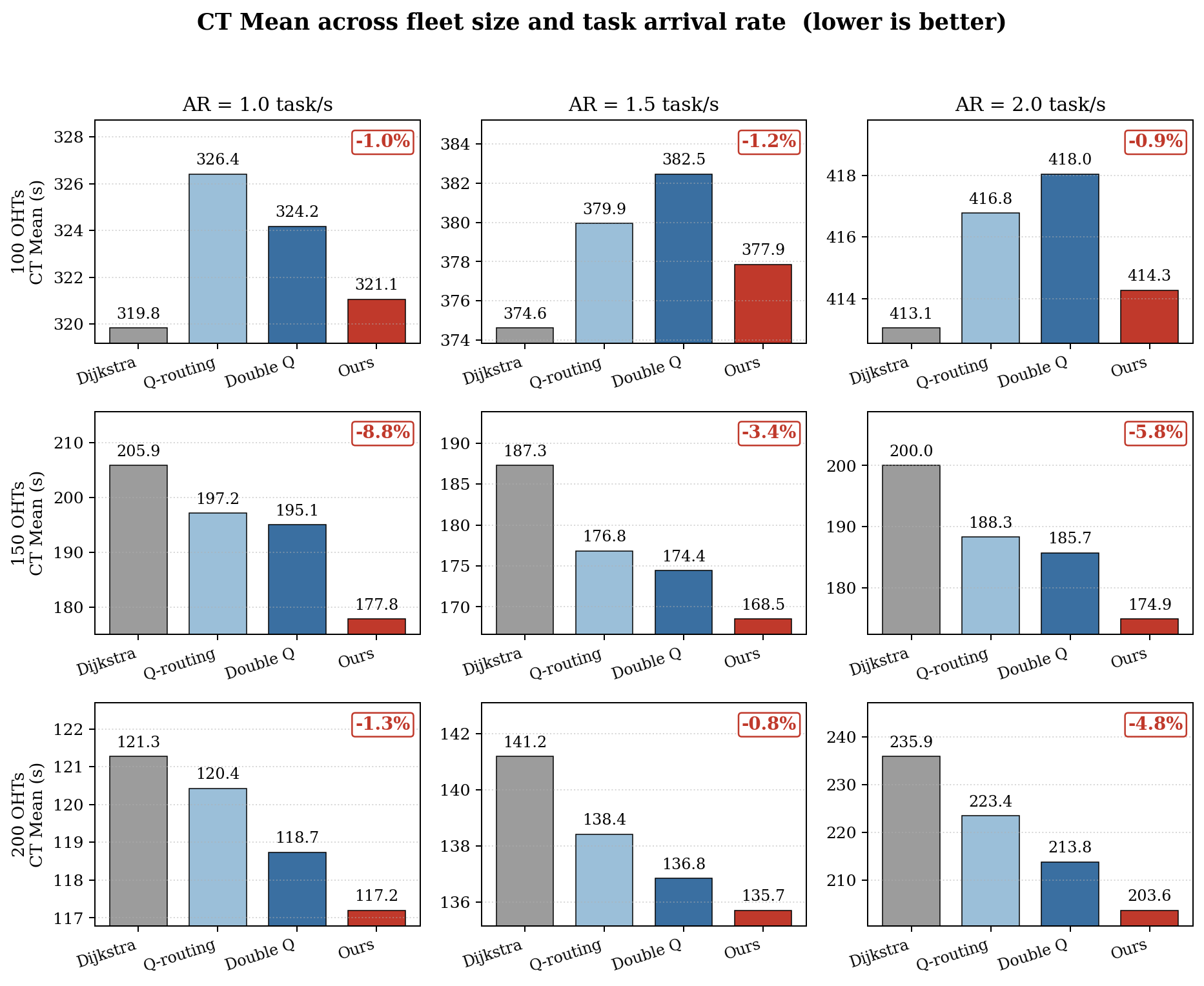}
\caption{CT Mean across the fleet-size and arrival-rate sweep, visualizing Table~\ref{tab:main_sweep}. Red annotations report the relative change of QNeuralDouble from tabular Double Q-routing; negative values indicate lower CT Mean.}
\label{fig:sweep_ct_mean}
\end{figure*}

Relative to QDouble, QNeuralDouble reduces CT Mean in every cell by $0.8\%$--$8.8\%$; completed-task counts stay within $1\%$ of the baseline in all but one cell. At 150\,OHTs the CT Mean reductions are $3.4\%$--$8.8\%$, and CT P95 improves by $3.4\%$--$6.2\%$. Fig.~\ref{fig:sweep_ct_mean} visualizes the full CT Mean sweep.

\subsection{Ablations}
\label{sec:ablation}

The warm-start study uses the 150- and 200-OHT scenarios in Table~\ref{tab:warmstart}; the component study uses the 150-OHT, AR$=1.0$ cell in Table~\ref{tab:ablation}.

\begin{figure}[t]
\centering
\begin{minipage}[t]{0.8\linewidth}
\centering
\includegraphics[width=0.8\linewidth]{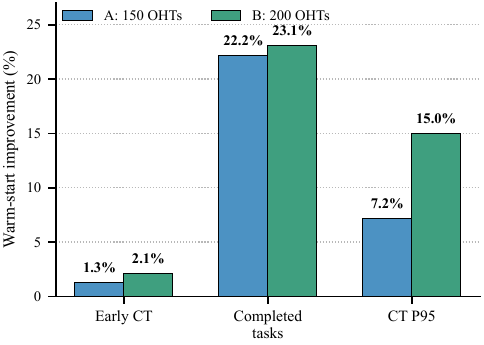}
\small (a)
\end{minipage}\hfill
\begin{minipage}[t]{0.8\linewidth}
\centering
\includegraphics[width=0.8\linewidth]{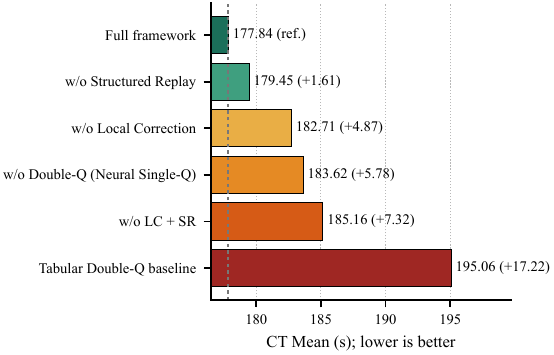}
\small (b)
\end{minipage}
\caption{Warm-start and component analyses. (a) Relative warm-start improvement over cold start during the first $200$\,s; Early CT is the mean CT of tasks completed in that interval. (b) CT Mean for the neural component deletions and the tabular Double-Q baseline; parentheses report the difference from the full framework. The panels visualize Tables~\ref{tab:warmstart} and~\ref{tab:ablation}, respectively.}
\label{fig:ablation_visuals}
\end{figure}

Table~\ref{tab:warmstart} compares warm-started and cold-started QNeuralDouble. Early CT, defined as the mean CT of tasks completed within the first $200$\,s, improves by 1.3\% (Scen.~A) and 2.1\% (Scen.~B). Over the same startup window, the warm-started router completes 22.2\% and 23.1\% more tasks, while CT P95 decreases by 7.2\% and 15.0\%. Fig.~\ref{fig:ablation_visuals}(a) visualizes these relative changes.

\begin{table}[t]
\centering
\caption{Warm-start comparison on matched startup scenes (first 200\,s, AR$=$1.0). Early CT is the mean CT of tasks completed within this interval. Values are mean $\pm$ std over 5 seeds.}
\label{tab:warmstart}
\scriptsize
\setlength{\tabcolsep}{4pt}
\begin{adjustbox}{max width=\columnwidth}
\begin{tabular}{llrrrrrr}
\toprule
Scen. & Variant & Early CT & Compl. & CT Mean & CT P95 & $\Delta$ Compl. & $\Delta$ P95 \\
\midrule
A (150) & Cold & 51.08 $\pm$ 1.43 & 27 $\pm$ 0.8 & 59.04 $\pm$ 1.20 & 92.36 $\pm$ 3.19 & (ref.) & (ref.) \\
A (150) & Warm & \textbf{50.42 $\pm$ 1.26} & \textbf{33 $\pm$ 1.8} & \textbf{57.62 $\pm$ 1.35} & \textbf{85.71 $\pm$ 3.08} & \textbf{+22.2\%} & \textbf{--7.2\%} \\
\midrule
B (200) & Cold & 54.17 $\pm$ 1.40 & 26 $\pm$ 1.4 & 61.93 $\pm$ 1.53 & 110.84 $\pm$ 3.83 & (ref.) & (ref.) \\
B (200) & Warm & \textbf{53.05 $\pm$ 1.18} & \textbf{32 $\pm$ 1.6} & \textbf{60.74 $\pm$ 1.28} & \textbf{94.18 $\pm$ 3.25} & \textbf{+23.1\%} & \textbf{--15.0\%} \\
\bottomrule
\end{tabular}
\end{adjustbox}
\end{table}

For the online-component study, we use the 150\,OHT, AR$=1.0$ cell. Starting from the full model, we separately disable structured replay (SR; unit loss weights and no event-stratified sampling), local correction (LC; $\lambda_t\!\equiv\!\beta_t\!\equiv\!0$), Double-Q (single estimator), and both LC$+$SR. Offline warm start and the matched-scene protocol are fixed for these neural component deletions. The final row is the ordinary tabular Double Q-routing baseline and is therefore a framework-level comparison rather than a single-component ablation. Table~\ref{tab:ablation} reports the results.

\begin{table}[t]
\centering
\caption{Component and baseline comparisons at 150\,OHT, AR$=1.0$ (mean over 5 seeds). $\Delta$ is the CT Mean difference from the full model (positive is worse). LC: local congestion correction. SR: structured replay.}
\label{tab:ablation}
\scriptsize
\begin{adjustbox}{max width=\columnwidth}
\begin{tabular}{lrrrr}
\toprule
Variant & CT Mean & Compl. & CT P95 & $\Delta$ CT Mean \\
\midrule
QNeuralDouble (Full) & \textbf{177.84} & 588 & \textbf{362.77} & (ref.) \\
$-$~w/o Structured Replay & 179.45 & 587 & 372.84 & +1.61 \\
$-$~w/o Local Correction & 182.71 & 585 & 384.06 & +4.87 \\
$-$~w/o Double-Q (Neural Single-Q) & 183.62 & 586 & 386.94 & +5.78 \\
$-$~w/o Both (LC + SR) & 185.16 & 586 & 379.42 & +7.32 \\
Tabular Double-Q baseline & 195.06 & 592 & 386.75 & +17.22 \\
\bottomrule
\end{tabular}
\end{adjustbox}
\end{table}

Among the neural component deletions, replacing Double-Q with a single estimator produces the largest mean CT penalty ($+5.78$\,s), followed by removing local correction ($+4.87$\,s) and structured replay ($+1.61$\,s). Removing LC and SR together gives a $+7.32$\,s penalty, so their observed effects are not additive in this cell. The full framework differs from the tabular Double-Q baseline by $17.22$\,s; because this comparison changes the representation together with the initialization and adaptation mechanisms supported by it, the difference is an overall framework comparison and is not attributed to the neural representation alone. Fig.~\ref{fig:ablation_visuals}(b) visualizes the CT Mean comparisons.

\section{Discussion and Limitations}
\label{sec:discussion}

In the selected ablation cell, each online component has a positive mean contribution, although the effects of local correction and structured replay are not additive (Table~\ref{tab:ablation}). The tabular Double-Q row changes the representation, initialization, and adaptation design together, so its larger gap measures the complete framework rather than the neural representation alone. The warm-start study addresses a different question: Early CT changes modestly, while completed-task count and CT P95 change more clearly during the first $200$\,s (Table~\ref{tab:warmstart}).

The method ranking changes with both fleet size and task arrival rate. Dijkstra obtains the lowest CT Mean in all three 100-OHT cells, whereas QNeuralDouble performs best in all 150- and 200-OHT cells. Relative to tabular Double Q-routing, the largest reductions occur at 150\,OHTs, ranging from 3.4\% to 8.8\%. At 200\,OHTs, the reductions are 1.3\%, 0.8\%, and 4.8\% at AR$=1.0$, 1.5, and 2.0, respectively, with the largest reduction at AR$=2.0$. The evidence therefore supports a setting-dependent benefit rather than a monotone fleet-size or arrival-rate effect. Because utilization and congestion-source statistics are not reported, we do not attribute these patterns to a specific capacity regime.

\textbf{Limitations.} The evaluation is on a single fab layout within one simulator and not validated on a physical OHT system. The external baselines are tabular policies; we include a neural single-Q variant as an internal ablation but do not compare against GNN routers (e.g., HarmonyRouting~\cite{kang2026harmonyrouting}, for which a public implementation is unavailable) or multi-agent RL baselines. The offline prior mixes three reasonable but suboptimal policies, and we have not characterized robustness to substantially worse or adversarial logged data. We fix the task-dispatching rule and study routing in isolation, though dispatching and routing interact. The learning criterion uses decision-epoch discounting and domain-weighted shaping. Future work will study duration-aware discounting, potential-based shaping, and sensitivity to the reward weights and deadlock-warning thresholds in Appendix~\ref{app:impl_reward}.



\section{Conclusion}
\label{sec:conclusion}

We presented a neural Double Q-routing framework for large-scale OHT systems. It replaces classical Q-routing's destination-indexed table with a shared state-action value function, initializes that function from mixed Dijkstra/Q/Double-Q trajectories, and refines it online with Double-Q updates, local congestion correction, and structured replay. Across matched simulations spanning 100/150/200 OHTs and 1.0/1.5/2.0\,task\,s$^{-1}$ arrival rates, the framework improves CT Mean over tabular Double Q-routing in every cell (up to 8.8\%), while completed-task counts remain within 1\% in eight of nine cells.

The ablations qualify this aggregate result. In the selected cell, each online component has a positive mean contribution, and the complete framework also outperforms the tabular Double-Q baseline. Offline initialization matters most during startup: it raises the number of tasks completed during the first $200$\,s by up to 23\% and reduces startup tail latency by up to 15\%.


\appendices

\section{Additional Analyses and Implementation Details}
\label{sec:appendix}
This appendix is organized from general analysis to reproducibility details. It first compares the computational complexity of tabular and neural Q-routing, then provides the complete state and action feature definitions and examines value-estimate stability. Finally, it records the implementation settings for offline pretraining, reward design, local correction, and replay storage.

\subsection{Complexity analysis}
\label{app:complexity}

\textbf{Space complexity.} The tabular Q-routing methods maintain a three-element tuple $(d,i,j)$ representing the destination, current node, and next hop. Let $|D|$ denote the number of distinct destinations, $|V|$ the number of nodes, and $|E|$ the number of directed edges. The table requires $\sum_{i \in V} |D| \cdot \deg^+(i) = |D| \cdot |E|$ scalar entries. In our fab guideway, $|V|=3684$, $|E|=4257$, and $|D|=608$ yield approximately $2.59$M independent parameters that must each be visited and updated separately. Double Q-routing doubles this to around $5.18$M entries.

Our neural value function replaces the entire table with a two-hidden-layer MLP. With input dimension $d_\phi$ (the joint state-action feature size), hidden size $h$, and a single scalar output, the parameter count is $d_\phi h+h+h^2+h+h+1=O(d_\phi h+h^2)$, including the three bias vectors. This count is independent of the destination set size $|D|$, the number of nodes $|V|$, and the number of directed edges $|E|$. With $d_\phi=24$ and $h=64$, the network has $1536+64+4096+64+64+1=5825$ parameters. This is about $445\times$ fewer parameters than the tabular baseline has entries and $889\times$ fewer than tabular Double Q. The neural model scales as $O(d_\phi h+h^2)$, whereas the tabular storage grows as $O(|D|\,|E|)$.

\textbf{Time complexity.} At each junction with $k$ feasible next hops, the tabular router performs $k$ table lookups at $O(1)$ each, giving $O(k)$ per decision. The neural router evaluates a forward pass for each candidate, costing $O(k(d_\phi h + h^2))$. In our guideway, the maximum out-degree is two as all split nodes are binary, so each decision requires at most two forward passes of about $5800$ multiply-adds.

\subsection{Full state and action feature definitions}
\label{app:features}

\textbf{State features ($d_s = 10$).} The state $s$ observed by a vehicle at a decision node $i$ with target $d$ contains: (i) the current and target node identifiers encoded as normalized scalar indices $\mathrm{idx}(i)/(|V|-1)$ and $\mathrm{idx}(d)/(|V|-1)$, which provide a coarse positional cue without inflating the input dimension with one-hot vectors over $|V|=3684$ nodes; (ii) the shortest-path distance $h(i,d)$, normalized by $D_s=\bar w\sqrt{|V|}$, where $\bar w$ is the mean edge weight; (iii) the in-degree and out-degree of $i$, each normalized by the maximum degree in $G$; (iv) a load/unload-node indicator that is $1$ when $i$ is associated with a task pickup or drop-off point and $0$ otherwise; (v) a one-hot encoding of the task phase (pickup, delivery, or other); and (vi) the recent waiting time of the vehicle, normalized by a fixed reference of \SI{5}{s}. All continuous components lie in $[0,1]$ after normalization, and all binary components take values in $\{0,1\}$, so that no single feature dominates the input scale.

\textbf{Action features ($d_a = 14$).} For each candidate next hop $j$, the action vector contains: (i) the normalized scalar index of $j$; (ii) the normalized length of edge $(i,j)$; (iii) the normalized remaining distance $h(j,d)$; (iv) the progress $h(i,d)-h(j,d)$, affinely mapped to $[0,1]$; (v) the normalized in- and out-degrees of $j$; (vi) the occupancy of $(i,j)$ relative to its length-based capacity; (vii) the waiting-queue length at $j$, normalized by the same fixed five-vehicle reference used by the implementation; and (viii) the following six pressure summaries.

\subsection{Value-estimate stability of tabular vs.\ neural Q-routing}
\label{app:param_sharing}

\begin{figure}[h]
\centering
\includegraphics[width=0.9\columnwidth]{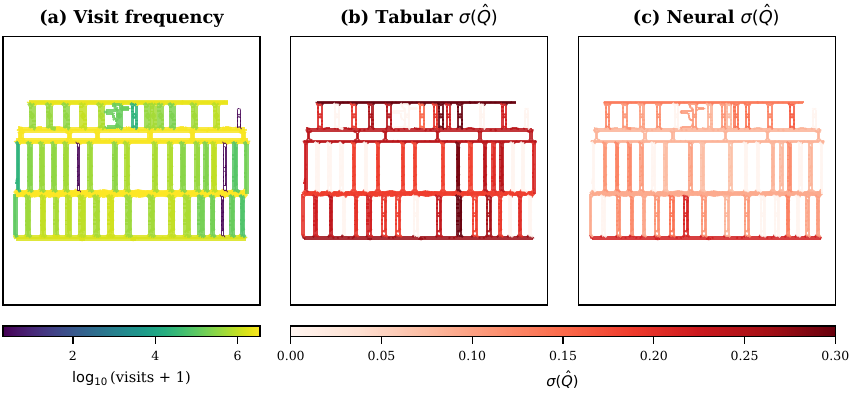}
\caption{Value-estimate stability of tabular and neural Q-routing on the fab guideway of Fig.~\ref{fig:fab_map}. All panels use the same layout, with color aggregated by bay. (a) Visit frequency of $(d,i,j)$ tuples. (b) Standard deviation of tabular estimates over the final snapshots. (c) Standard deviation of neural estimates over the same snapshots.}
\label{fig:param_sharing}
\end{figure}

For both routers, $\sigma(\hat{Q})$ measures how much a value estimate fluctuates near convergence. During the final training window, we take periodic snapshots of $\hat{Q}(d,i,j)$ for every tuple and report the standard deviation over the last few snapshots. Fig.~\ref{fig:param_sharing} shows higher tabular variation across most of the guideway and lower variation for the shared neural value function. This pattern is consistent with parameter sharing across routing contexts, but the figure measures stability rather than value accuracy; routing outcomes are reported in Section~\ref{sec:experiments}.

\subsection{Implementation Details}
\label{app:impl}

The reported configuration is held fixed across the fleet-by-arrival-rate
sweep.

\subsubsection{Offline pretraining and dataset construction}
\label{app:impl_offline}

\textbf{Dataset composition.} The offline data come from the three behavior
policies and three collection seeds specified in Section~\ref{sec:method}.
Strata indexed by operating cell, policy, and seed are sampled equally so that
larger strata do not dominate through sample count. Return-to-go accumulation
uses the boundaries in Eq.~\eqref{eq:rtg}. The online general and event replay
stores contain 5,000 and 2,000 transitions, respectively, and both discard the
oldest record when full.

\subsubsection{Reward design and weights}
\label{app:impl_reward}

The reward in Eq.~\eqref{eq:reward} has three groups of terms: the per-edge
realized-time penalties summed over $e\in\mathcal I_t$, the split-level local
risk and route-progress terms, and the per-edge warning penalty. The time terms
set the principal scale, with separate weights for motion, waiting, and
blocking. Risk and progress are smaller anticipatory terms in the same training
criterion.

The weight tuple is reported after Eq.~\eqref{eq:reward}; movement time defines
the reference scale, waiting and blocking remain on the same scale, and the
risk and progress terms are smaller shaping terms. The warning proxy takes the
three values $0$, $1$, and $3$. Warning is triggered when waiting exceeds
$3\,\mathrm{s}$ with an immediate downstream queue of at least two; it becomes
critical after the warning condition persists for $5\,\mathrm{s}$. These
settings are fixed for every experimental cell.

\subsubsection{Local correction and replay storage}
\label{app:impl_local}

The local residual in Eq.~\eqref{eq:local_residual_update} remains in the same
units as the base value. The count term in Eq.~\eqref{eq:local_correction} is a
decaying pessimism heuristic rather than a calibrated confidence bound. Replay
labels, target shares, loss multipliers, precedence, and sparse-class fallback
are specified in Section~\ref{sec:online_adaptation} and
Table~\ref{tab:replay_labels}.

\balance
\bibliographystyle{IEEEtran}
\bibliography{ref}

\end{document}